%% file: main.tex
\documentclass[10pt,twocolumn,letterpaper]{article}

\usepackage{cvpr}              
\usepackage{booktabs}
\usepackage{booktabs,tabularx,makecell,array,pifont}

\usepackage{pifont}
\newcommand{\cmark}{\ding{51}} 
\newcommand{\xmark}{\ding{55}} 
\usepackage{siunitx}
\newcolumntype{Y}{>{\centering\arraybackslash}X}
\usepackage{booktabs,tabularx,makecell,array,pifont}
\newcolumntype{M}[1]{>{\centering\arraybackslash}m{#1}}
\usepackage{booktabs}
\usepackage{siunitx}
\usepackage[table]{xcolor}
\usepackage{bbm}
\usepackage{multirow}
\usepackage{makecell}
\usepackage{makecell} 
\usepackage{xspace}
\usepackage{graphics} 
\usepackage{epsfig} 
\usepackage{amsmath} 
\usepackage{amssymb}  

\usepackage{bbding}
\usepackage{gensymb}
\usepackage[utf8]{inputenc} 
\usepackage[T1]{fontenc}    
\usepackage{url}            
\usepackage{booktabs}       
\usepackage{amsfonts}       
\usepackage{nicefrac}       
\usepackage{microtype}      
\usepackage{color}
\usepackage[table]{xcolor}
\usepackage{caption}
\usepackage{graphicx}
\usepackage{float}
\usepackage{multirow}
\usepackage{diagbox}

\usepackage{algorithm}
\usepackage{algpseudocode}
\usepackage{arydshln}
\usepackage{colortbl}

\usepackage{multicol}
\usepackage{multirow}
\usepackage{listings}
\input{preamble}

\definecolor{cvprblue}{rgb}{0.21,0.49,0.74}
\usepackage[pagebackref,breaklinks,colorlinks,allcolors=cvprblue]{hyperref}

\title{EgoTL: Egocentric Think-Aloud Chains for Long-Horizon Tasks}

\author{
\begin{tabular}{c}
Lulin Liu\textsuperscript{1,2}\footnotemark[1] \quad
Dayou Li\textsuperscript{2}\footnotemark[1] \quad
Yiqing Liang\textsuperscript{3} \quad
Sicong Jiang\textsuperscript{4,5} \quad
Hitesh Vijay\textsuperscript{2}  \\
Hezhen Hu\textsuperscript{6} \quad
Xuhai Xu\textsuperscript{7} \quad
Zirui Liu\textsuperscript{1} \quad
Srinivas Shakkottai\textsuperscript{2} \quad
Manling Li\textsuperscript{8} \quad
Zhiwen Fan\textsuperscript{2}\textsuperscript{$\dagger$} \\
\textsuperscript{1}UMN \quad
\textsuperscript{2}TAMU \quad
\textsuperscript{3}Brown University \quad
\textsuperscript{4}McGill University \\
\textsuperscript{5}2077AI \quad
\textsuperscript{6}UT Austin \quad
\textsuperscript{7}Columbia University \quad
\textsuperscript{8}Northwestern University \\
\\
{\textcolor[HTML]{497BB8}{\textbf{Project Website:}} \href{https://ego-tl.github.io/}{\textcolor[HTML]{497BB8}{https://ego-tl.github.io/}}}
\end{tabular}
\vspace{0.0cm}
}
\begin{document}

\twocolumn[{%
\renewcommand\twocolumn[1][]{#1}%


\maketitle
\begin{center}
    \centering
    \captionsetup{type=figure}
    \vspace{-8mm}
  \includegraphics[width=\textwidth]{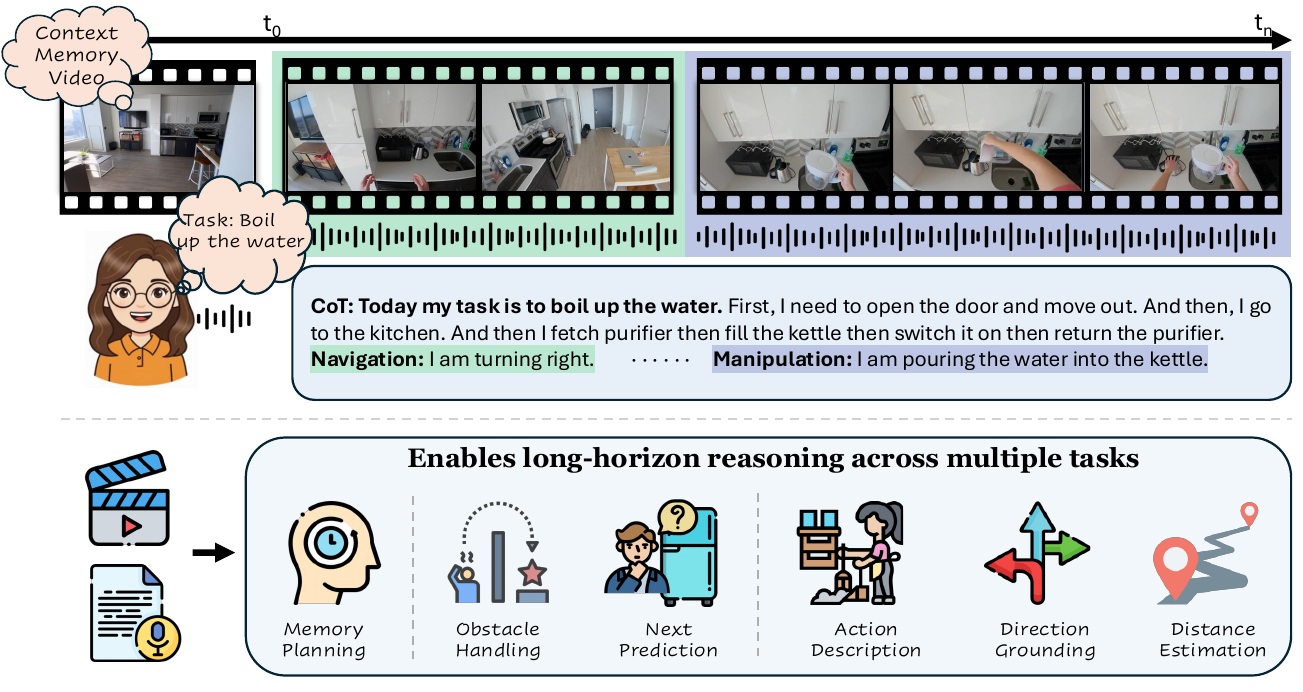}
  \vspace{-6mm}
\captionof{figure}{What is the right data for teaching current vision-language foundation models human-like spatial perception and long-horizon egocentric reasoning? EgoTL introduces a say-before-act capture pipeline that records abstract household goals, think-aloud chains of thought, and explicit navigation and manipulation steps before execution. Grounded in metric 3D reconstructions and explicit action labels, EgoTL enables human-aligned supervision and diagnosis for long-horizon egocentric spatial reasoning.}
    \label{fig:teaser}
\end{center}%
}]

\renewcommand{\thefootnote}{\fnsymbol{footnote}}
\footnotetext[1]{Equal contribution, Lulin Liu was a visiting student at Texas A\&M University during this work.}

\input{sec/0_abstract}    
\input{sec/1_intro}

\input{sec/2_related_w}
\input{sec/3_our_data}

\input{sec/4_data_annotation}
\input{sec/4_eval}

\input{sec/6_con}

{
    \small
    \bibliographystyle{ieeenat_fullname}
    \bibliography{main}
}
\input{sec/X_suppl}

\end{document}

%% file: preamble.tex
\usepackage[table]{xcolor}
\definecolor{navyblue}{HTML}{0071BC}
\definecolor{hotpink}{HTML}{FF0080}

\definecolor{oai-white}{HTML}{FFFFFF}
\definecolor{oai-black}{HTML}{000000}
\definecolor{oai-red}{HTML}{FF4500}
\definecolor{oai-green}{HTML}{51DA4C}
\definecolor{oai-blue}{HTML}{0000FF}
\definecolor{oai-yellow}{HTML}{FFF639}
\definecolor{oai-magenta}{HTML}{FF45FF}
\definecolor{oai-cyan}{HTML}{00FFFF}
\definecolor{oai-orange}{HTML}{FE7600}
\definecolor{oai-violet}{HTML}{8A2BE2}
\definecolor{oai-brown}{HTML}{A0522D}
\definecolor{oai-green-050}{HTML}{F4FFF4}
\definecolor{oai-green-100}{HTML}{E9FFE8}
\definecolor{oai-green-200}{HTML}{D9FFD8}
\definecolor{oai-green-300}{HTML}{C9FFC7}
\definecolor{oai-green-400}{HTML}{A6FFA3}
\definecolor{oai-green-500}{HTML}{7CF178}
\definecolor{oai-green-600}{HTML}{51DA4C}
\definecolor{oai-green-700}{HTML}{3FA93B}
\definecolor{oai-green-800}{HTML}{2D712A}
\definecolor{oai-green-900}{HTML}{193718}
\definecolor{oai-gray-000}{HTML}{FFFFFF}
\definecolor{oai-gray-100}{HTML}{FAFAFA}
\definecolor{oai-gray-200}{HTML}{F5F5F5}
\definecolor{oai-gray-300}{HTML}{E5E5E5}
\definecolor{oai-gray-400}{HTML}{FFB7A4}
\definecolor{oai-gray-500}{HTML}{CDCDCD}
\definecolor{oai-gray-600}{HTML}{A8A8A8}
\definecolor{oai-gray-700}{HTML}{747474}
\definecolor{oai-gray-800}{HTML}{393939}
\definecolor{oai-gray-900}{HTML}{000000}



%% file: sec/0_abstract.tex
\begin{abstract}
Large foundation models have made significant advances in embodied intelligence, enabling synthesis and reasoning over egocentric input for household tasks. However, VLM-based auto-labeling is often noisy because the primary data sources lack accurate human action labels, chain-of-thought (CoT), and spatial annotations; these errors are amplified during long-horizon spatial instruction following. These issues stem from insufficient coverage of minute-long, daily household planning tasks and from inaccurate spatial grounding. As a result, VLM reasoning chains and world-model synthesis can hallucinate objects, skip steps, or fail to respect real-world physical attributes. To address these gaps, we introduce \textbf{EgoTL}. EgoTL builds a think-aloud capture pipeline for egocentric data. It uses a say-before-act protocol to record step-by-step goals and spoken reasoning with word-level timestamps, then calibrates physical properties with metric-scale spatial estimators, a memory-bank walkthrough for scene context, and clip-level tags for navigation instructions and detailed manipulation actions. With EgoTL, we are able to benchmark VLMs and World Models on six task dimensions from three layers and long-horizon generation over minute-long sequences across over 100 daily household tasks. We find that foundation models still fall short as egocentric assistants or open-world simulators. Finally, we finetune foundation models with human CoT aligned with metric labels on the training split of EgoTL, which improves long-horizon planning and reasoning, step-wise reasoning, instruction following, and spatial grounding.

\end{abstract}

%% file: sec/1_intro.tex
\section{Introduction}
\label{sec:intro}
\begin{figure*}[t]
\centering
\vspace{-5mm}
        \includegraphics[clip,trim=0cm 0cm 0cm 0cm,width=1.0\linewidth]{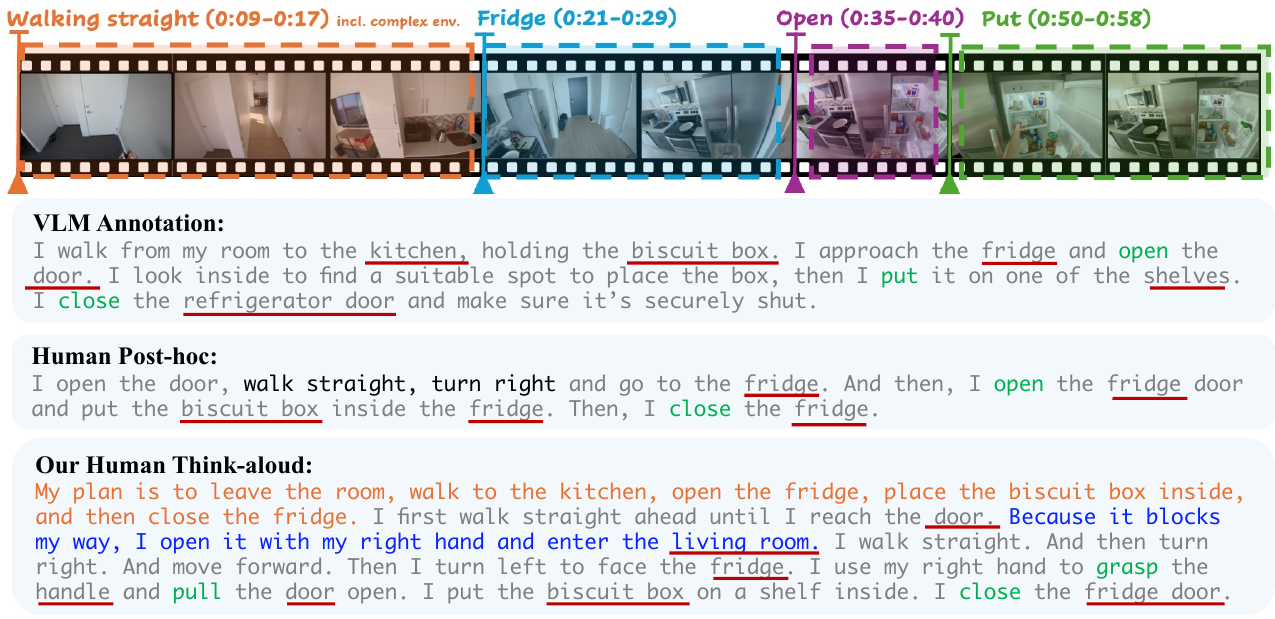}
\vspace{-7mm}
     \caption{A comparison of video annotations for an everyday task. The top filmstrip shows keyframes of the task "put a biscuit box in the fridge," segmented into four sub-actions. Below, we compare three types of textual descriptions: (1) VLM annotation from Qwen2.5-VL 32B \cite{qwen2.5}, (2) a \textbf{Human Post-hoc} description created after the event, and (3) our proposed \textbf{Human Think-aloud} transcript with Say-Before-Act protocol, captured in real-time. The think-aloud data provides a richer, more detailed description of the actions and intentions. To analyze the richness of the think-aloud data, we apply semantic highlighting based on specific conditions: \textbf{Green} is used for verbs (e.g., `walk`, `open`) to mark explicit actions. Nouns are \textbf{underscored in red} (e.g., \underline{fridge}) only when the object they refer to is visible in the corresponding video frame. \textbf{Orange} highlights the user's internal "chain of thought", revealing high-level planning and reasoning. Finally, \textbf{blue} marks "scene-aware" descriptions, indicating the user's spatial awareness beyond immediate object interaction.}
    \label{fig:task_annotation}
    \vspace{-4mm}
\end{figure*}

Large-scale foundation models \cite{qwen2,agarwal2025cosmos, wang2025internvl3_5, wan2025, naveed2024comprehensiveoverviewlargelanguage, wei2022emergent, beguvs2023large, zhang-etal-2024-unveiling-linguistic, li2024learning,radford2018improving,radford2019language,brown2020language,touvron2023llama,touvron2023llama2,bai2023qwen,team2023gemini} have significantly advanced embodied intelligence by enabling agents to learn from human data and reason over egocentric input~\cite{bai2023qwen}, as well as to synthesize future states as open-world simulators~\cite{wan2025}. This progress is driven by web-scale knowledge transferred to egocentric agents, allowing them to understand, reason, and plan before acting.
However, training these models relies on massive amounts of real-world data, and primary data sources (e.g., web video) lack accurate human action labels, chain-of-thought (CoT), and spatial annotations. The problem is further amplified during day-to-day household, minute-long long-horizon spatial instruction following, because many automatic annotation pipelines for both benchmark and web egocentric videos produce temporally misaligned captions \cite{pei2025egothinker} and provide insufficient coverage of minute-long, day-to-day household planning tasks, leading to weak planning, spatial grounding, and causal reasoning under long-horizon goals in complex environments.
Another line of work reduces reliance on automatic annotation pipelines by using human annotation with explicit timing, as in Ego4D \cite{grauman2022ego4d} and HD-EPIC \cite{perrett2025hd}. These datasets provide short video clips (a few seconds) with human-verified labels for actions and hand-object interactions, along with temporal boundaries and object references. However, because most labels are written post hoc, they are not tightly aligned to timestamps and rarely capture reasoning that spans multiple clips. As a result, descriptions emphasize local segments rather than the stepwise plan that links them. Moreover, many existing collections are video-only rather than truly multimodal, which limits disambiguation of user intent and state transitions in first-person views. In Figure~\ref{fig:task_annotation}, we show the differences between the annotation methods. By incorporating synchronized audio, it becomes possible to record spoken cues and on-the-fly explanations, providing a more faithful account of the actor's real-time reasoning.

To close this gap, we raise the question: \textit{can we build collection-time annotations with step-by-step subgoals and provide calibrated multimodal datasets that align spoken reasoning, actions, and spatial context at capture time?} In this paper, we present EgoTL, a long-horizon multimodal dataset covering a broad set of household tasks. Each sequence unifies detailed navigation steps and manipulation actions under explicit task goals and human chains of thought. We propose a say-before-act protocol to record every intermediate goal and spoken reasoning with word-level timestamps, calibrating physical properties with metric-scale spatial estimators. This approach captures a critical, often-missing intention signal. For example, our think-aloud protocol might capture: ``I was going to walk straight to the object, but the chair blocks the path, so I will move the chair first, then continue.'' Post-hoc narration typically collapses this nuance into: ``I moved the chair and continued forward.'' This highlights a theory-of-mind \cite{jin2024mmtom} gap: a subject's intention (why-now, why-this) is not directly observable, yet long-horizon tasks critically depend on it for replanning. By recording intention before execution, Say-Before-Act yields data that is more precise, time-aligned to the upcoming action, and less outcome-conditioned than post-hoc narration.

As shown in Figure~\ref{fig:teaser}, EgoTL starts with an overall task description, followed by multiple action episodes. Each episode contains synchronized video and audio collected via this think-aloud protocol, where the operator states the next goal before execution and explains key steps during navigation and manipulation. This produces time-aligned, word-level chains of thought paired with metric-scale spatial labels and a memory-bank walkthrough detailing object locations and room layouts for later planning. We then segment each long video into single-navigation or manipulation clips, adding clip-level tags such as manipulation descriptions, walking distances in meters, and turning directions.

Building on EgoTL, we introduce a benchmark, EgoTL-Bench, that probes vision-language models (VLMs) and world models (WMs). Specifically, we evaluate planning and spatial grounding across six dimensions on more than 100 household tasks. Testing against strong VLM baselines reveals common failure modes: skipping key steps, drifting over time, and failing on spatially grounded questions. We also evaluate WMs by simulating realistic human-object interactions within complex but typical home layouts, where current models struggle to follow instructions and maintain object persistence and metric consistency across long rollouts. Finally, using our training split, we fine-tune existing foundation models on EgoTL. We demonstrate that aligning spoken human CoT with metric-scale labels improves long-horizon planning and long-video rollouts, yielding better reasoning and more consistent generation performance.
\begin{itemize}
\item We identify limitations of existing large foundation models on household common tasks and find typical failure patterns: skipped steps, object hallucinations, temporal drift, and weak spatial reasoning, which we link to the post-hoc annotation and VLM auto-labeling paradigm.

\item We introduce EgoTL, a think-aloud protocol for recording egocentric video with synchronized audio. It uses a say-before-act protocol to log step-by-step goals and reasoning, capturing the intentions and producing human-aligned chains of thought with word-level timestamps.

\item EgoTL further calibrates physical properties with metric-scale spatial estimators, maintains a memory-bank walkthrough for scene context, and adds clip-level tags for navigation instructions and detailed manipulation actions.

\item Built on over 100 household tasks with long-horizon CoT, we evaluate current VLMs and world models across six dimensions spanning three layers, revealing systematic errors in planning, temporal alignment, and metric consistency. Furthermore, fine-tuning these models on the training split yields significantly improved long-horizon rollouts, fewer skipped steps, and stronger instruction following and spatial grounding.
\end{itemize}

%% file: sec/2_related_w.tex
\section{Related Work}

\paragraph{Egocentric Video Dataset.}
Recent egocentric video datasets range from large-scale human-captured collections to synthetic environments with fine-grained annotations. Early egocentric data were mainly collected from the Internet; Assembly101~\cite{sener2022assembly101} contains 4,321 videos of human-object interactions with detailed action labels. Large efforts such as Ego4D~\cite{grauman2022ego4d} and Ego-Exo4D~\cite{grauman2024ego} provide over 3,000 hours of egocentric video of household activities and diverse real-world scenarios. With the emergence of Aria glasses~\cite{engel2023project}, many newer datasets focus on everyday human-object interactions, including Aria Everyday Objects~\cite{straub24efm}, Aria Everyday Activities~\cite{lv2024aria}, and Nymeria~\cite{ma2024nymeria}, whose wide field-of-view cameras better capture attention and rich visual context. Recent work emphasizes high-quality, precise annotations rather than just recording length: Ego4D~\cite{grauman2022ego4d} and HD-EPIC~\cite{perrett2025hd} rely on dense human post-hoc descriptions, while datasets such as HOT3D~\cite{banerjee2025hot3d} and HOI-4D~\cite{liu2022hoi4d} use multiple sensors to obtain ground-truth signals for downstream tasks. In Table \ref{tab:datasets_overview}, we show the statistics of the existing dataset. Complementing these real-world datasets, the synthetic ALFRED dataset~\cite{shridhar2020alfred} integrates navigation and manipulation in a single environment with precise action and navigation labels under fully controlled conditions.

\vspace{-3mm}
\paragraph{Large Foundation Models.}
Large-scale foundation models~\cite{qwen2, qwen2.5, chen2024internvl2, lin2024vila, fu2025video} have made rapid progress in embodied intelligence, enabling agents to reason over egocentric input and synthesize future states. Early multimodal VLMs such as Qwen and InternVL achieve strong performance on standard video understanding benchmarks, showing that large-scale pre-training supports accurate recognition and question answering on short clips. Long-video VLMs, including LongVLM and Long-VILA, extend this to minutes of video, integrating information across events for complex procedures, while world models and video generators such as COSMOS~\cite{agarwal2025cosmos} and WAN~\cite{wan2025} treat video synthesis as environment simulation, predicting long-horizon rollouts from actions or language prompts~\cite{agarwal2025cosmos,wan2025,zhang2025world,team2025hunyuanworld}. More recently, this line of work has shifted toward spatial intelligence: benchmarks such as VSI-Bench~\cite{yang2024thinking}, VSTI-Bench~\cite{fan2025vlm}, SpatialBench~\cite{cai2025spatialbot}, and MindCube~\cite{yin2025spatial} ask models to reason about egocentric directions, relative distances, object locations, and 3D-consistent layouts. These capabilities are crucial for embodied agents that must follow long-horizon instructions, maintain object permanence, and plan navigation and manipulation in real homes. However, most egocentric datasets label long-horizon videos using automatic or post-hoc tools, without recording human actions and reasoning before execution, which introduces temporal drift, missing steps, and spatially inconsistent supervision. These limitations motivate EgoTL, which pairs think-aloud egocentric supervision with metric spatial calibration to provide human-aligned labels for long-horizon egocentric reasoning.

\begin{table*}[t]
\centering
\setlength{\tabcolsep}{4pt}
\footnotesize
\begin{tabular}{c c c c c c c c c}
\toprule
Dataset
& Year
& Narration
& CoT
& Task
& Audio
& Nav.\ Ann.
& Reasoning.
& Capturing Devices \\
\midrule
EPIC.\text{-}100 \cite{Damen2020Collection}
  & 2021 & post-hoc     & \xmark & \cmark & \cmark & \xmark    & \xmark   & Headworn          \\
Ego4D~\cite{grauman2022ego4d}
  & 2022 & post-hoc     & \xmark & \xmark   & \cmark & \xmark    & \xmark   & GoPro/Vuzix/PupilLabs \\
HOI4D~\cite{liu2022hoi4d} 
  & 2022 & -            & \xmark & \xmark & \xmark & \xmark   & \xmark    & Helmet+RGB\!-\!D      \\
Aria. Obj. \cite{straub24efm}
  & 2023 & -            & \xmark & \xmark & \xmark & \xmark   & \xmark    & Aria                  \\
Ego\text{-}Exo4D \cite{grauman2024ego}
  & 2023 & post-hoc     & \xmark & 43     & \cmark & \xmark  & \xmark   & Aria\,+\,exocams      \\
Holo\text{-}Assist \cite{wang2023holoassist}
  & 2023 & real time    & \xmark & \cmark & \cmark & \xmark & \xmark & HoloLens2             \\
ARCTIC~\cite{fan2023arctic} 
  & 2023 & -            & \xmark & \xmark & \xmark & \xmark & \xmark & Helmet                \\
EgoVid \cite{wang2024egovid}
  & 2024 & post-hoc     & \xmark & \xmark & \xmark & \xmark & \xmark & Various               \\
Aria. Act. \cite{lv2024aria}
  & 2024 & real time    & \xmark & \xmark & \cmark & \xmark & \xmark & Aria                  \\
Nymeria~\cite{ma2024nymeria}
  & 2024 & post-hoc     & \xmark & \xmark & \cmark & \xmark & \xmark & Aria                  \\
HO\text{-}Cap~\cite{wang2024ho} 
  & 2024 & -            & \xmark & \xmark & \xmark & \xmark & \xmark & HoloLens\,+\,RGB\!-\!D \\
HOT3D \cite{banerjee2025hot3d}
  & 2025 & -            & \xmark & \xmark & \cmark & \xmark & \xmark & Aria/Quest3           \\
HD\text{-}EPIC~\cite{perrett2025hd}
  & 2025 & post-hoc            & \xmark & \xmark & \cmark & \xmark & \xmark & Headworn          \\
EgoMe \cite{qiu2025egome}
  & 2025 & -            & \xmark & \cmark & \cmark & \xmark & \xmark & Headworn          \\
\rowcolor{gray!15}
\textbf{EgoTL (ours)}
  & 2025 & think-aloud  & \cmark & >100    & \cmark & \cmark & \cmark & Aria/Headworn/Wayfarer \\
\bottomrule
\end{tabular}
\caption{\textbf{Egocentric datasets with reasoning annotations.} 
Comparison of egocentric video datasets relevant to long-horizon, task-oriented navigation and manipulation. 
Columns indicate whether each dataset provides narration (\emph{Narration}, including post-hoc or real-time “Think-Aloud”), explicit chain-of-thought supervision (\emph{CoT}), task labels (\emph{Task}), synchronized audio (\emph{Audio}), navigation annotations (\emph{Nav.\ Ann.}), and reasoning traces under complex cases (\emph{Reason.}).
\cmark/\xmark denote presence/absence; “Think-Aloud” denotes real-time narration and reasoning during task execution. 
Rows are ordered by year of release, with EgoTL (ours) shown last.}
\label{tab:datasets_overview}
\end{table*}

%% file: sec/3_our_data.tex
\section{EgoTL Collection Principles}
\label{sec:dataset}

We introduce EgoTL, an egocentric, minute-long, task-oriented, multimodal video dataset that makes explicit how humans plan and act along with intentions when given an abstract goal. Each recorded unit is an \emph{episode}. Formally, we denote an episode by
\begin{equation}
   E = (M, R, \mathcal{A}, \mathcal{C}),
\end{equation}
where $M$ is a memory-bank walkthrough video, $R$ is the task audio-video recording, $\mathcal{A}$ is the episode-level chain of thought (CoT), and $\mathcal{C}$ is the set of clip-level annotations. Intuitively, $M$ captures the spatial context, while $R$ captures the language-conditioned execution. Thus, each episode provides episode-level reasoning ($\mathcal{A}$) and fine-grained, clip-level supervision ($\mathcal{C}$) that covers navigation distance, turning direction, and manipulation description.

\paragraph{Datasets Statistics.}
The dataset spans 400 episodes across more than 100 tasks with 2-4 minutes each, and each episode consists of annotated metadata and videos. We captured the dataset using Meta Aria research glasses~\cite{engel2023project}, Meta Ray-Ban Wayfarer glasses, and smartphones mounted on a head strap.

\begin{figure}[t]
\centering
        \includegraphics[clip,trim=0cm 0cm 0cm 0cm,width=0.9\linewidth]{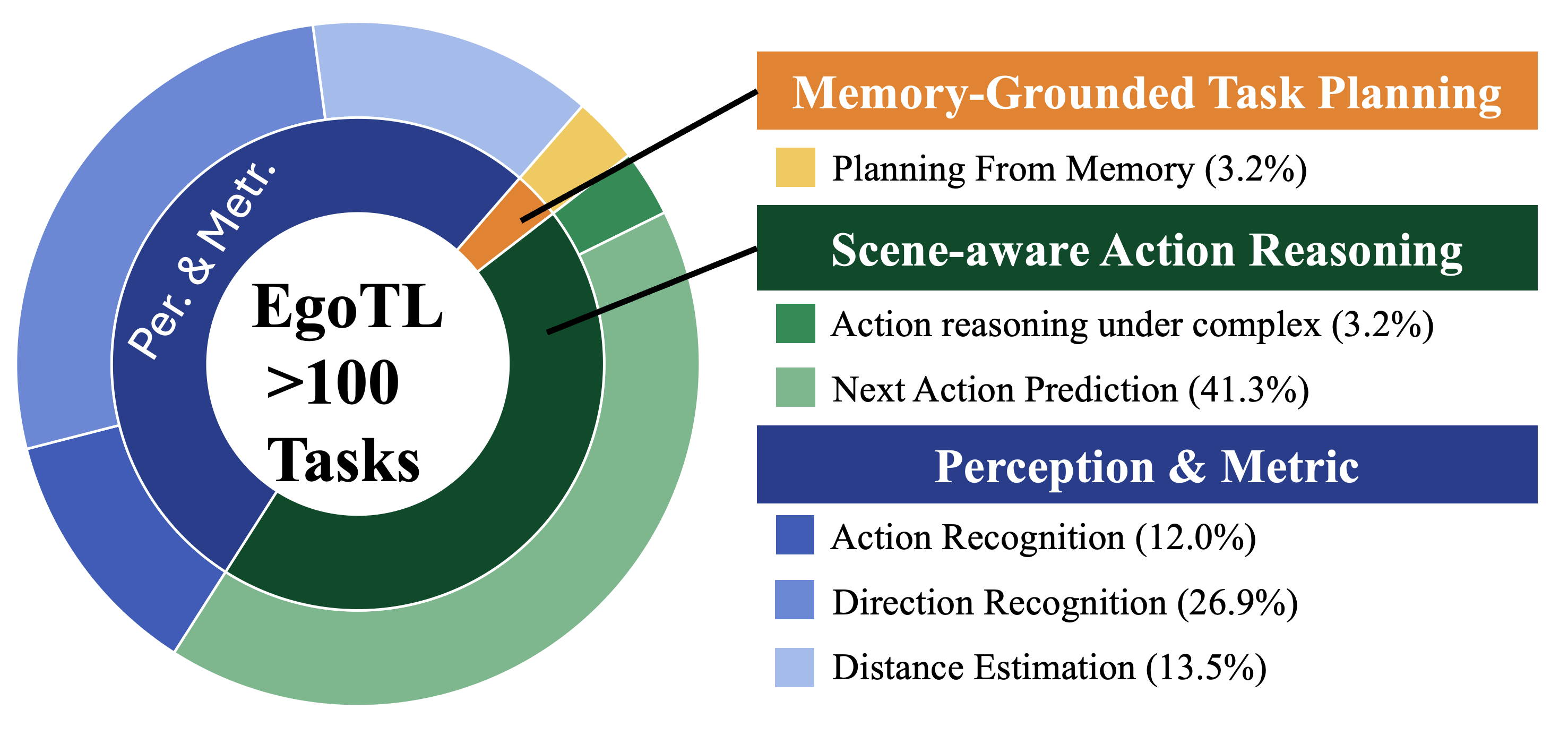}
    \vspace{-3mm}
    \caption{\textbf{Benchmark statistics.} Distribution of EgoTL benchmark tasks across three main categories.}
    \label{fig:dataset-overview}
    \vspace{-6mm}
\end{figure}

\vspace{-3mm}
\paragraph{Participants and Recruitment.}
We recruited 50 volunteers from the University of Minnesota, Twin Cities, and Texas A\&M University. Our study protocol was reviewed and approved by UMN and TAMU Institutional Review Board (IRB). Participants were recruited through online postings and flyers and received modest compensation. All participants completed a brief English proficiency screening, and only those who passed were enrolled. Before recording, participants were trained on device usage (wearing and operating head-mounted cameras), safety considerations (avoiding hazards during walking and manipulation), and the think-aloud requirement. During training, they practiced the ``say-before-act'' protocol on a small set of example tasks until they could consistently verbalize their intentions and actions in complete, understandable sentences.

\vspace{-3mm}
\paragraph{Chain of Thought (CoT) Collection.}
For each episode \(E_i\), the subject first articulates the abstract task (e.g., ``My task is to get a bottle of milk''), and then verbalizes the episode-level chain of thought \(\mathcal{A}_i\) to decompose this goal into ordered subtasks (e.g., ``I need to go to the kitchen, open the fridge, and pick up the milk''). The subject then specifies the current location (e.g., ``I am in the bedroom''). After that, the subject follows the ``say-before-act'' protocol: for navigation, the subject must state the intended motion (``I am walking straight,'' ``I am turning right,'' ``I am turning left'') and only then perform it; for manipulation, upon reaching the target location, the subject must state the manipulation in the same manner (e.g., ``I am closing the lid,'' ``I am pulling the fridge handle with my right hand'') and then execute it. Each such spoken-and-executed unit is stored in the clip-level annotations \(\mathcal{C}_i\) defined above, where each element consists of the narrated intent and its corresponding structured description (navigation distance, turning direction, or manipulation details).
We also instruct participants to introduce at least one unexpected scene obstruction during the episode that is not included in the initial episode-level CoT \(\mathcal{A}_i\). For example, when the task is ``take the milk from the fridge,'' the participant may place a bag in front of the fridge door, so they must first verbalize and perform a clearance action before resuming the original plan; this additional step is added as another element in \(\mathcal{C}_i\). This design allows the dataset to capture language-conditioned recovery behaviors in the presence of occlusions or layout-induced constraints, which are typically missed by VLM-only annotations.

\vspace{-3mm}
\paragraph{Memory Bank.}

To provide richer spatial context beyond the task execution itself, we record a memory-bank walkthrough for each episode. Lasting approximately two minutes, this contextual video is captured once all task episodes conclude and the environment is systematically restored to its initial state (e.g., doors closed, objects returned, and receptacles reset). The annotator uses this walkthrough to document the overarching layout of the involved rooms and the interiors of specific storage spaces. The resulting memory-bank video, denoted as $M$, is temporally aligned with the corresponding task recording $R$ from the same episode to provide spatial context for training and evaluation.

%% file: sec/4_data_annotation.tex
\section{EgoTL Data Curation Pipeline}
\label{sec:curation}

After data collection, we convert the raw egocentric audio-video streams into structured supervision aligned at both the episode and clip levels. Our curation pipeline consists of two main stages: (i) speech-to-text alignment, which extracts human chain-of-thought (CoT) and execution descriptions with word-level precision, and (ii) clip-wise segmentation, which assigns navigation and manipulation labels, turning direction, and metric-scale walking distance.
\begin{table}[t]
\centering
\vspace{-3mm}
\resizebox{0.7\linewidth}{!}{%
\begin{tabular}{p{0.26\linewidth} p{0.66\linewidth}}
\hline
\textbf{Aspect} & \textbf{Description} \\
\hline
CoT occurrence & CoT appears at the episode start (plan announcement) and at moments where the subject revises the plan due to unexpected obstacles or layout changes. \\
CoT storage & The corresponding CoT segment is attached to the first clip where it becomes relevant and is also stored as episode-level metadata. \\
Per-clip video & Video segment corresponding to the utterance-aligned time span. \\
Per-clip transcript & Exact human transcript aligned at the word level. \\
Per-clip label & Clip category label: \textit{navigation} or \textit{manipulation}, with navigation sub-types (e.g., \textit{walk-straight}, \textit{turn-left}). \\
Contextual CoT & Pointer to the associated episode-level CoT, enabling models to condition on both local action descriptions and global task-level reasoning. \\
\hline
\end{tabular}%
}
\caption{Clip-level CoT occurrence and stored metadata in EgoTL.}
\label{tab:cot_clip_level}
\end{table}

\vspace{-3mm}
\paragraph{Speech-to-Text Alignment.}
Every episode is recorded under the \textit{think-aloud} and \textit{say-before-act} protocol, meaning that subjects verbalize both their high-level task plan and every navigation or manipulation action immediately before performing it. Consequently, the audio stream already contains ground-truth human reasoning and action phrases without requiring post-hoc reconstruction. We transcribe each episode using WhisperX~\cite{bain2022whisperx}, which provides word-level start and end timestamps and yields a fully time-aligned transcript. This allows us to precisely anchor (i) abstract task descriptions, (ii) episode-level CoT (the stepwise ``I need to \dots'' reasoning), (iii) navigation statements (e.g., ``I am walking straight,'' ``I am turning left''), and (iv) manipulation statements (e.g., ``I am closing the lid'') to the video timeline. Because the CoT originates from human speech immediately before execution, our reasoning traces capture real-time intent rather than post-hoc explanations.
\begin{figure*}[t]
\centering
\vspace{-5mm}
        \includegraphics[clip,trim=0cm 0cm 0cm 0cm,width=1.0\linewidth]{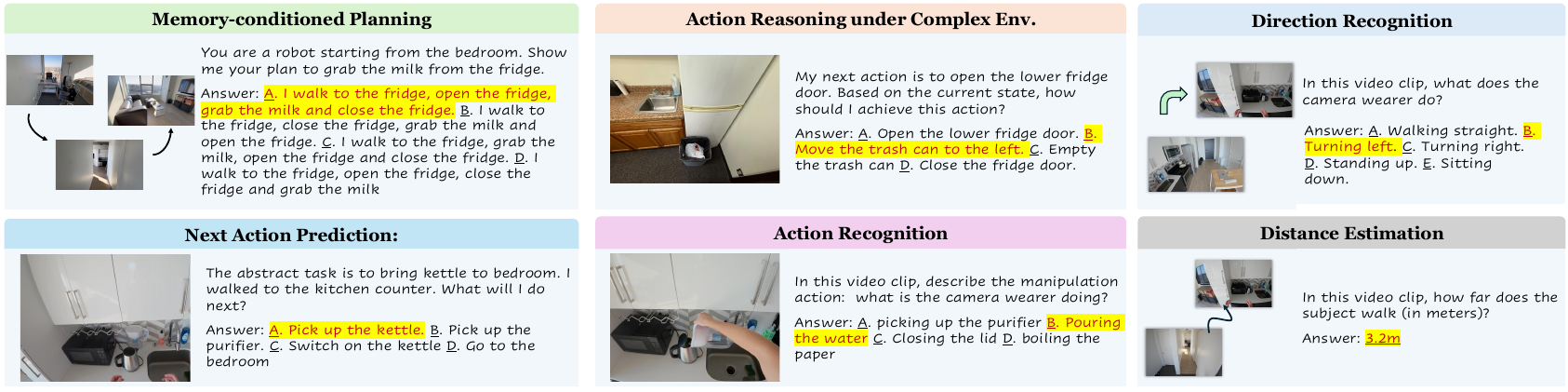}
\vspace{-7mm}
\caption{\textbf{Task overview in EgoTL-Bench.} EgoTL-Bench decomposes egocentric spatial understanding into six tasks across three layers. \emph{Memory-conditioned planning} asks the model to generate an action plan from a memory-bank walkthrough and a high-level goal. \emph{Scene-aware action reasoning} tests whether it selects the correct action in cluttered scenes, such as moving an obstacle before opening a door. \emph{Next action prediction} checks if the model can infer the immediate next step from the current frame and abstract task. At the perceptual layer, \emph{action recognition} describes the ongoing manipulation, \emph{direction recognition} identifies egocentric motion primitives (walking straight, turning, standing or sitting), and \emph{distance estimation} predicts how far the subject walks in meters. The shown examples are simplified; the full benchmark uses more templates, distractors, and longer episodes.}
    \label{fig:task_demonstration}
    \vspace{-4mm}
\end{figure*}
\vspace{-3mm}
\paragraph{Clip-Wise Segmentation and Annotation.}

We segment each long-horizon episode into short clips using aligned timestamps. A silence gap longer than 2 seconds indicates a new clip boundary, and the corresponding video span with its transcribed utterance is treated as one clip. Each clip is assigned to one of two high-level categories, \textit{navigation} or \textit{manipulation}. Navigation clips are further sub-typed into five atomic motion primitives: \textit{walk-straight}, \textit{turn-left}, \textit{turn-right}, \textit{standing-up}, and \textit{sitting-down}. These primitives cover both horizontal and vertical egomotion and enable finer-grained evaluation of VLMs on 3D spatial understanding.
\vspace{-3mm}
\paragraph{Categorization Strategy.}
Categorization is performed \emph{directly from the spoken content}. For each clip, we examine the utterance immediately preceding the action:
\begin{itemize}
    \item If the utterance matches one of the commands (e.g., ``I am walking straight,'' ``I am turning left,'' ``I am turning right''), we assign the corresponding navigation label.
    \item If the utterance contains a verb-object manipulation phrase (e.g., ``I am closing the lid'', ``I am opening the drawer''), the clip is marked as \textit{manipulation}.
    \item If a clip contains CoT-like narration rather than an action command, and no navigation verb appears, it is treated as non-action narration or merged into the preceding clip.
\end{itemize}
This strategy keeps the clip semantics tightly coupled to human-intended commands.

\vspace{-3mm}
\paragraph{CoT Presence at Clip Level.}
Only a subset of clips contain explicit CoT. CoT typically appears (i) at the beginning of the episode when subjects announce their plan, and (ii) when unexpected events in complex environments require plan revision. When this happens, the CoT segment is stored in the first relevant clip and also preserved as episode-level metadata. For every clip, we maintain both local action information and global task context, as summarized in Table~\ref{tab:cot_clip_level}.

\vspace{-3mm}
\paragraph{Walking Distance Estimation.}
For clips labeled \textit{walk-straight}, we compute the metric-scale travel distance using MapAnything~\cite{keetha2025mapanything}. We uniformly sample frames at 10 fps and obtain a metric-scale camera center for each frame:
\begin{equation}
\mathbf{p}_i = (x_i, y_i, z_i) \in \mathbb{R}^3.
\end{equation}
Given the ordered sequence $\{\mathbf{p}_0, \ldots, \mathbf{p}_N\}$, we define the traveled distance as
\begin{equation}
L = \sum_{i=1}^{N} \left\| \mathbf{p}_i - \mathbf{p}_{i-1} \right\|_2.
\end{equation}
This provides a physically meaningful walking distance for every \textit{walk-straight} navigation clip in our dataset and serves as metric supervision for egocentric spatial reasoning.

%% file: sec/4_eval.tex
\section{Benchmarking Large Foundation Models \\ under EgoTL}
\label{sec:evaluation}
\begin{table*}[ht!]
    \captionsetup{type=table}
    \centering
    \begin{minipage}{1.0\textwidth}
    \centering
    \fontsize{4.6pt}{4.4pt}\selectfont
    \setlength\tabcolsep{3pt}
    \renewcommand{\arraystretch}{1.2}
    \resizebox{0.85\textwidth}{!}{
    \begin{tabular}{r|cc|cccccc} 
    & & &
    \rotatebox{75}{Memory-Cond.\ Plan} &
    \rotatebox{75}{Scene-Aware Interact.} &
    \rotatebox{75}{Next-Action Pred.} &
    \rotatebox{75}{Action Recog.} &
    \rotatebox{75}{Direction Recog.} &
    \rotatebox{75}{Distance Est.\ (MRA)} \\
    Methods & Rank & Avg. &
    \multicolumn{5}{c}{\cellcolor{yellow!10}Multiple-Choice Answer} &
    \multicolumn{1}{c}{\cellcolor{orange!10}Numerical Answer} \\
    \hline

    \rowcolor{navyblue!5}
    \multicolumn{9}{l}{\textcolor{black}{\textit{Chance Level Baselines}}} \\
    Chance Level (Random)
    & - & 0.25 & 0.25 & 0.25 & 0.25 & 0.25 & 0.25 & -- \\
    Chance Level (Frequency)
    & - & 0.36 & 0.38 & 0.31 & 0.42 & 0.40 & 0.29 & -- \\
    \hline

    \rowcolor{navyblue!5}
    \multicolumn{9}{l}{\textcolor{black}{\textit{Open-source VLMs}}} \\
    Qwen2.5-VL 7B
    & 4 & 0.4672
    & 0.4773 & 0.3202 & 0.4803 & 0.5628 & 0.4954
    & \cellcolor{oai-gray-600}\textbf{20.04\%} \\
    Qwen2.5-VL 32B
    & 3 & 0.5218
    & \cellcolor{oai-gray-600}\textbf{0.6136} & 0.3801 & 0.4253 & 0.6104 & \cellcolor{oai-gray-300}{0.5798}
    & \cellcolor{oai-gray-300}{13.85\%} \\
    InternVL 2.5 8B
    & 6 & 0.4181
    & 0.2955 & 0.2912 & 0.4779 & 0.6753 & 0.3505
    & 7.96\% \\
    InternVL 2.5 38B
    & \cellcolor{oai-green-400}{2} & \cellcolor{oai-gray-300}{0.5508}
    & 0.5682 & \cellcolor{oai-gray-300}{0.4301} & \cellcolor{oai-gray-300}{0.5137} & \cellcolor{oai-gray-600}\textbf{0.7099} & 0.5321
    & 1.35\% \\
    InternVL 3 8B
    & 5 & 0.4525
    & 0.5000 & 0.3121 & 0.4301 & 0.5671 & 0.4532
    & 4.71\% \\
    InternVL 3 38B
    & \cellcolor{oai-green-600}{1} & \cellcolor{oai-gray-600}\textbf{0.5808}
    & \cellcolor{oai-gray-300}{0.5909} & \cellcolor{oai-gray-600}\textbf{0.4318} & \cellcolor{oai-gray-600}\textbf{0.5281} & \cellcolor{oai-gray-300}{0.6797} & \cellcolor{oai-gray-600}\textbf{0.6734}
    & 3.07\% \\
    \hline

    \rowcolor{navyblue!5}
    \multicolumn{9}{l}{\textcolor{black}{\textit{Proprietary VLMs}}} \\
    GPT-5~\cite{achiam2023gpt}
    & \cellcolor{oai-green-600}{1} & 0.5660
    & 0.6086 & 0.6786 & 0.3231 & 0.5542 & 0.6653
    & 3.76\% \\
    GPT-4o~\cite{hurst2024gpt}
    & 4 & 0.4270
    & 0.4347 & 0.3571 & 0.3365 & 0.4416 & 0.5651
    & 10.15\% \\
    Gemini2.0-Flash~\cite{team2023gemini}
    & \cellcolor{oai-green-200}{3} & 0.4455
    & 0.3478 & 0.3750 & 0.4313 & 0.5238 & 0.5498
    & 33.83\% \\
    Gemini2.5-Flash~\cite{comanici2025gemini}
    & \cellcolor{oai-green-400}{2} & 0.5120
    & 0.4024 & 0.4156 & 0.4931 & 0.5913 & 0.6574
    & 16.28\% \\
    \hline

    \rowcolor{navyblue!5}
    \multicolumn{9}{l}{\textcolor{black}{\textit{Finetuned Model on EgoTL}}} \\
    \textbf{Ours}
    & \cellcolor{oai-green-600}{1} & \textbf{0.6826}
    & \textbf{0.6956} & \textbf{0.7500} & \textbf{0.5707} & \textbf{0.7125} & \textbf{0.6843}
    & \textbf{39.45\%} \\
    \hline
    \end{tabular}
    }
    \caption{\textbf{Evaluation of VLMs on EgoTL.}
    For each task within the open-source VLMs group, \colorbox{oai-gray-600}{dark gray} highlights the best open-source model and \colorbox{oai-gray-300}{light gray} denotes the second-best open-source model.}
    \label{tab:egocot_main}
    \end{minipage}
    \vspace{-0.1cm}
    \vspace{-3mm}
\end{table*}
We introduce EgoTL-Bench to quantitatively evaluate large foundation models on long-horizon spatial reasoning from egocentric videos. EgoTL-Bench covers diverse model families, parameter scales, and training recipes. While existing benchmarks such as VSI-Bench~\cite{yang2025thinking} and VSTI-Bench~\cite{fan2025vlm} provide valuable evaluation of spatial intelligence, they do not include ground-truth reasoning traces aligned with execution, which are crucial for planning and error localization. Under EgoTL-Bench, we evaluate open-source VLMs, including Qwen2.5-VL~\cite{qwen2.5}, InternVL2.5~\cite{chen2024internvl2}, and InternVL3~\cite{zhu2025internvl3}, as well as proprietary VLMs such as Gemini 2.0 Flash~\cite{team2023gemini} and Gemini 2.5 Flash~\cite{comanici2025gemini}. In Sec.~\ref{sec:evaluation}, we evaluate video world models on long-horizon rollout fidelity.

\vspace{-4mm}
\paragraph{Task Definition.}
Many VLMs already perform well on short-horizon, single-scene, weakly constrained benchmarks such as VSI-Bench~\cite{yang2025thinking} and VSTI-Bench~\cite{fan2025vlm}. In contrast, EgoTL-Bench focuses on long-horizon egocentric video and is designed to probe how models perceive and reason about spatial information over time. As illustrated in Figure~\ref{fig:task_demonstration}, we decompose evaluation into three layers so that we can localize failure modes instead of reporting only a single aggregate score. The distribution of questions across these tasks is shown in Figure~\ref{fig:dataset-overview}.

\begin{itemize}
    \item \textbf{Top layer: Memory-conditioned planning.} Models must plan from stored memory-bank videos, selecting a feasible sequence of waypoints or actions to complete the task.
    \item \textbf{Middle layer: Scene-aware action reasoning.} This layer is split into (i) action reasoning under complex environments, where models must infer what the human is doing given clutter and occlusions, and (ii) next-action prediction, where models must predict the next step in the sequence from the current egocentric view and context.
    \item \textbf{Bottom layer: Perceptual and metric reasoning.} Here we evaluate human action recognition, direction recognition, and distance estimation. Unlike prior benchmarks that focus mainly on horizontal directions (left, right, straight), EgoTL-Bench also includes vertical motions such as standing up and sitting down, provide a more complete test of 3D egocentric spatial understanding.
\end{itemize}

\vspace{-4mm}
\paragraph{Question-Answer Generation.}
Question-answer (QA) pairs are primarily generated automatically by combining human think-aloud annotations with predefined question templates (see appendix for full templates). 


\vspace{-4mm}
\paragraph{Metrics.}
For multiple-choice answer (MCA) tasks, we use standard \textit{accuracy} ($\mathcal{ACC}$)~\cite{fu2025video,hendrycks2020measuring,yue2024mmmu} based on exact matching (optionally with fuzzy matching for minor wording differences). For numerical-answer (NA) tasks, we use mean relative accuracy (MRA)~\cite{yang2025thinking}, which measures prediction quality across multiple tolerance levels. Accuracy at a single threshold only reflects relative error within a narrow band, so $\mathcal{MRA}$ instead averages relative accuracy over a set of confidence thresholds $\mathcal{C} = \{0.5, 0.55, \dots, 0.95\}$:
\vspace{-0.2cm}
\begin{equation}
\vspace{-0.2cm}
\begin{small}
\mathcal{MRA} = \frac{1}{|\mathcal{C}|} \sum_{\theta \in \mathcal{C}} \mathbbm{1} \left( \frac{|\hat{y} - y|}{y} < 1 - \theta \right),
\end{small}
\end{equation}
where $y$ and $\hat{y}$ denote the ground-truth and predicted values, respectively. This metric rewards predictions that stay within a small relative error across a wide range of tolerances.

\begin{table*}[t]
\centering
\setlength{\tabcolsep}{4pt}
\footnotesize

\sisetup{
  table-number-alignment = center,
  table-format = 2.2,
  round-mode = places,
  round-precision = 2,
  detect-weight = true
}
\vspace{-3mm}
\resizebox{0.85\textwidth}{!}{%
\begin{tabular}{l *{6}{S} *{3}{c}}
\toprule
\multirow{2}{*}{\textbf{Method}} &
\multicolumn{6}{c}{CLIP Score $\uparrow$} &
\multicolumn{3}{c}{VBench $\uparrow$} \\
\cmidrule(lr){2-7}\cmidrule(lr){8-10}
& \multicolumn{1}{c}{0--10\,s} & \multicolumn{1}{c}{10--20\,s} &
\multicolumn{1}{c}{20--30\,s} & \multicolumn{1}{c}{30--40\,s} &
\multicolumn{1}{c}{40--50\,s} & \multicolumn{1}{c}{50--60\,s} &
\multicolumn{1}{c}{\begin{tabular}[c]{@{}c@{}}Image\\Quality $\uparrow$\end{tabular}} &
\multicolumn{1}{c}{\begin{tabular}[c]{@{}c@{}}Subject\\Consistency $\uparrow$\end{tabular}} &
\multicolumn{1}{c}{\begin{tabular}[c]{@{}c@{}}Background\\Consistency $\uparrow$\end{tabular}}  \\
\midrule
COSMOS (vanilla)
  & \textbf{23.04} & \textbf{21.99} & 21.90 & 22.21 & \textbf{21.15} & 20.86
  & 0.58 & \textbf{0.80} & 0.86 \\
WAN (vanilla)
  & 22.11 & 20.94 & 20.20 & 19.68 & 20.07 & 19.81
  & 0.71 & 0.78 & 0.82 \\
\textbf{COSMOS (w/ EgoTL)}
  & 21.76 & 21.54 & \textbf{22.71} & \textbf{22.35} & 21.01 & \textbf{23.02}
  & \textbf{0.71} & 0.79 & \textbf{0.88} \\
\bottomrule
\end{tabular}%
}
\caption{\textbf{Interactive long-horizon video evaluation on EgoTL.} Each evaluated world model generates a 60-second egocentric rollout conditioned on identical CoT sequences. \emph{CLIP Score} columns report text-video alignment across consecutive 10-second intervals (0--60s), where higher values indicate stronger prompt adherence. \emph{VBench} columns assess overall image quality, subject consistency, and background stability. Comparing off-the-shelf baselines (vanilla COSMOS and WAN) against our EgoTL-finetuned COSMOS highlights the substantial benefit of egocentric think-aloud supervision for robust long-horizon generation.}
\label{tab:interactive}
\vspace{-3mm}
\end{table*}
\vspace{-4mm}
\paragraph{Chance-Level Baselines.}
To contextualize model performance, we report two chance-level baselines:
\begin{itemize}
    \item \textit{Chance level (random).} For MCA tasks, this is the accuracy obtained by uniformly random selection among answer options. It is not applicable to NA tasks.
    \item \textit{Chance level (frequency).} For each task, this is the accuracy of a heuristic that selects the most frequent answer in the training distribution. This baseline indicates how much of a model's gain could be explained by exploiting answer frequency or class imbalance, rather than reasoning.
\end{itemize}

\vspace{-4mm}
\paragraph{Main Results.}
Table~\ref{tab:egocot_main} and \ref{tab:interactive} present comprehensive evaluation results on EgoTL-Bench. Below, we detail observations regarding open-source and closed-source VLMs, alongside both VLMs and world models fine-tuned on our dataset.

\vspace{-4mm}
\paragraph{Open-Source VLMs.}
On EgoTL-Bench, open-source VLMs already outperform chance by a large margin across all discrete tasks in Table~\ref{tab:egocot_main}, but they still fall short of human performance on long-horizon reasoning. Within this family, scaling generally improves performance: Qwen2.5-VL 32B substantially boosts memory-conditioned planning accuracy over its 7B counterpart, and InternVL~3 38B consistently outperforms the 8B variant on scene-aware interaction, next-action prediction, and direction recognition. Architectures exhibit distinct specializations: InternVL~3 38B achieves the best overall average accuracy and strongest mid-level reasoning, while InternVL~2.5 38B attains the highest action recognition score. Interestingly, the smaller Qwen2.5-VL 7B performs best on distance estimation, indicating that metric distance understanding remains unstable and is not simply resolved by scaling up model size. Compared with proprietary systems, the strongest open-source models are competitive or slightly superior on several perceptual-layer tasks, suggesting community models already provide a strong foundation for embodied benchmarks like EgoTL.

\vspace{-4mm}
\paragraph{Proprietary VLMs.}
We evaluate several closed-source VLMs, including GPT-5~\cite{achiam2023gpt}, GPT-4o~\cite{hurst2024gpt}, Gemini 2.0 Flash~\cite{team2023gemini}, and Gemini 2.5 Flash~\cite{comanici2025gemini}, comparing them with open-source baselines. On high-level tasks such as memory-conditioned planning and scene-aware action reasoning, all models remain far below human performance, indicating long-horizon planning is still challenging. Within this low absolute regime, closed-source systems consistently achieve higher scores than open-source models, suggesting a relative advantage in high-level reasoning. At the perceptual layer, however, this gap narrows or even reverses: strong open-source models are often comparable to, or slightly stronger than, closed-source ones on next-action prediction and action recognition. For direction recognition, both open-source and closed-source models exhibit a strong bias toward predicting ``Move forward,'' making it difficult to reliably distinguish turning motions. Moreover, almost all models perform poorly on distance estimation, indicating current VLMs still lack robust egocentric distance understanding. Overall, these results show that substantial progress is required to reach human-level egocentric spatial understanding.

\vspace{-4mm}
\paragraph{Finetuning Qwen on EgoTL.}
We fine-tune Qwen2.5-VL-7B-Instruct \cite{qwen2.5}, a 7B-parameter multimodal transformer with a frozen vision encoder and cross-modal adapters. To adapt the model to EgoTL without overfitting or incurring the full cost of dense fine-tuning, we apply low-rank adaptation (LoRA) \cite{hu2022lora} to the language backbone. By inserting rank-16 LoRA adapters into all transformer blocks while keeping the vision tower and multimodal projector frozen, the model specializes to EgoTL's spatial reasoning distribution while preserving its general-purpose capabilities. We fine-tune on a disjoint EgoTL subset that yields approximately 1.2k curated Q\&A pairs, and evaluate on a test set of 100 task videos spanning 15 scenes. As shown in Table \ref{tab:egocot_main}, our model surpasses the strongest baselines across all metrics, demonstrating notable gains in high-level planning and low-level perception. Particularly for distance estimation, where current VLMs universally struggle. Our model attains substantially higher mean relative accuracy, nearly doubling the MRA of the best pre-fine-tuning configuration. These improvements confirm that our human-annotated dataset provides VLMs with implicit scale calibration and reliable supervision, substantially advancing spatial reasoning.

\vspace{-4mm}
\paragraph{Benchmarking WMs and Finetuning COSMOS on EgoTL.}
Additionally, to test whether EgoTL improves long-horizon video prediction, we finetune the COSMOS world model on a subset of about 600 single-navigation or single-manipulation clips with fine-grained annotations. We apply low-rank adaptation (LoRA, rank 16) \cite{hu2022lora} to COSMOS-Predict2~\cite{agarwal2025cosmos} for 2k iterations and compare three models: our finetuned COSMOS-Predict2, the original COSMOS 2B~\cite{agarwal2025cosmos}, and WAN 2.2~\cite{wan2025}.

We follow a rollout strategy similar to prior world-model work~\cite{hafner2024dreamerv3,ha2018world}: given an episode and its ground-truth CoT, the model generates long-horizon visual rollouts conditioned on the planned actions, and we evaluate a fixed number of clips per episode. For each rollout, we compute CLIPScore~\cite{hessel2021clipscore} for semantic alignment and use VBench~\cite{huang2024vbench} for video quality and task success. As shown in Table~\ref{tab:interactive}, finetuning on EgoTL consistently increases CLIPScore and key VBench metrics, yields trajectories that better follow CoT instructions, and maintains object identity and layout over longer horizons. This indicates that EgoTL provides a more realistic and diverse training signal than single-scene datasets such as HD-EPIC~\cite{perrett2025hd}, and combining think-aloud CoT with metric spatial supervision benefits long-horizon world modeling.

%% file: sec/6_con.tex
\vspace{-1mm}
\section{Conclusion}

Current egocentric annotation pipelines remain a key bottleneck for training foundation models: VLM auto-labeling is noisy, and post-hoc descriptions are temporally misaligned and fail to capture intents. To address this, we introduce EgoTL, a multimodal dataset utilizing a say-before-act protocol that records abstract goals, think-aloud reasoning, and explicit navigation and manipulation steps prior to execution. Grounded in metric 3D and action labels, EgoTL enables human-aligned supervision for long-horizon reasoning. Across 400 episodes and >100 tasks, EgoTL-Bench reveals VLMs and world models struggle with planning, distance grounding, and long-horizon consistency. While fine-tuning on EgoTL improves planning, reasoning, and rollout coherence, a substantial gap to human performance remains.

\section{Acknowledgement}
This research has been supported by computing support on the Vista GPU Cluster through the Center for Generative AI (CGAI) and the Texas Advanced Computing Center (TACC) at the University of Texas at Austin, and the Research Stabilization Fund from Columbia University. We thank the Project Aria team for supporting this work by providing Aria glasses used in our research, and Meta Reality Labs for the gift funding. We also thank all contributors and partners whose efforts made the EgoTL dataset possible. Finally, we thank Nuo Chen and Bangya Liu for their valuable feedback on this project.

%% file: sec/X_suppl.tex
\clearpage

\setcounter{page}{1}
\maketitlesupplementary
\renewcommand\thesection{\Alph{section}}

\noindent This supplement is organized as follows:
\begin{itemize}[itemsep=0em]
\item Section~\ref{sec:benchmark-construction} details the benchmark construction pipeline, including dataset curation, Q\&A construction, clip-wise labeling, and 3D trajectory--based distance annotation. Section \ref{sec:qabenchmark-construction} lists the details about the Q\&A construction. Section~\ref{sec:supp-prompts} lists the full prompt templates used for all six evaluation tasks, covering memory-grounded planning, action reasoning, next-action prediction, action recognition, direction recognition, and distance estimation.
\item Section~\ref{sec:supp-closed-source} describes the evaluation setup for closed-source VLMs and how we adapt our benchmark to their interfaces and input constraints.
\item Section~\ref{sec:supp-vlm-finetune} provides VLM fine-tuning specifications, including model architectures and checkpoints (Section~\ref{sec:supp-arch}), training data configuration (Section~\ref{sec:supp-train-data}) and evaluation (Section \ref{sec:eval-lora}).
\end{itemize}



\section{Benchmark Construction Details}
\label{sec:benchmark-construction}


\subsection{Q\&A Construction}
\label{sec:qabenchmark-construction}

For this step, we use Gemini 2.5 Flash \cite{comanici2025gemini} to automatically synthesize three distractor options for each multiple-choice question. We first sample 100 representative Q\&A pairs from our tasks and use them as a pilot set. For these pairs, we prompt Gemini 2.5 Flash \cite{comanici2025gemini} to generate three distractor options per question. Human annotators then carefully review the generated options to evaluate their quality: they check that each distractor is (1) clearly incorrect yet still plausible given the question context, (2) linguistically clear and unambiguous, and (3) free of hallucinated content or information that cannot be inferred from the provided input. Based on this manual review, we iteratively refine the prompting strategy (e.g., by specifying stricter constraints on correctness, relevance, and style of the options) until annotators are satisfied that the model reliably produces high-quality distractors on the pilot set. Once this prompt design is stabilized, we apply the same prompting pipeline to all Q\&A pairs in our tasks to generate three distractor options for each question at scale. Each prompt is tailored to its corresponding task in order to match the task-specific context and reasoning requirements.

\begin{table*}[t]
    \renewcommand{\arraystretch}{1.5} 
    
    \begin{tabular}{p{3.5cm}|p{10cm}|p{3cm}}
        \textbf{Task} & \textbf{Question Template} & \textbf{Input} \\
        \hline
        Planning from Memory & 
        \textit{You are given an egocentric task \textcolor{red}{\{abstract\_task\}} and four candidate chains-of-thought (\textcolor{red}{A}, \textcolor{red}{B}, \textcolor{red}{C}, \textcolor{red}{D}). Options are shuffled; you MUST read and compare all options and assign each a score between 0 and 1 according to how well it matches the task and the memory-bank video \textcolor{red}{\{memory\_bank\}}. Then select the single best option and return ONLY a JSON object of the form \texttt{\{``scores'': \{``A'': 0.0, ``B'': 0.0, ``C'': 0.0, ``D'': 0.0\}, ``best'': ``A''\}}.} 
        & \textit{Memory-bank video + text prompt}
         \\
        \hline
        Action Reasoning under Complex Environment & 
        \textit{The image shows the current condition and you are given an egocentric \textcolor{red}{\{abstract\_task\}}. You are an egocentric action reasoning agent: based on the scene, decide what the camera wearer should do next while considering obstacles, free space, and object locations. You are given four candidate next actions (A--D). Carefully read the task and analyze the image, then select the single option that best accomplishes the task and is physically feasible in the scene. Output \textbf{only} the text of the chosen option, exactly as written, with no additional explanation or formatting.} 
        & \textit{Current frame + abstract task}
        \\
        \hline
        Next Action Recognition & 
        \textit{You are an egocentric next-action predictor. You will see the current egocentric video clip and a chain-of-thought describing the ongoing task \textcolor{red}{\{CoT\}}. From the candidate next actions (\textcolor{red}{\{option 1\}}, \textcolor{red}{\{option 2\}}, \textcolor{red}{\{option 3\}}, \textcolor{red}{\{option 4\}}), choose \textbf{exactly one} next action and output only the chosen option text.} 
        & \textit{Current video clip + CoT text}
        \\
        \hline
        Action Recognition & 
        \textit{You are an egocentric video action classifier. You will be given a short egocentric video clip and four candidate descriptions of the action, each labeled with a letter: \textcolor{red}{A}, \textcolor{red}{B}, \textcolor{red}{C}, and \textcolor{red}{D}. Select the \textbf{one} option that best matches the action shown in the video and answer with exactly one capital letter (A, B, C, or D) and nothing else. The four options are: \textcolor{red}{\{option 1\}}, \textcolor{red}{\{option 2\}}, \textcolor{red}{\{option 3\}}, \textcolor{red}{\{option 4\}}.} 
        & \textit{Current video clip} \\
        \hline
        Direction Recognition & 
       \textit{You are a video grounding agent. From the egocentric video, choose the dominant motion direction: \textcolor{red}{Turn left}, \textcolor{red}{Turn right}, \textcolor{red}{Move forward}, \textcolor{red}{Going up}, or \textcolor{red}{Going down}. Decide based on the \textbf{global} motion trend and ignore small local jitters. Output only the \textbf{text} of the chosen option (not the letter).} 
        & \textit{Current video clip} \\
        \hline
        Distance Estimation & 
        \textit{You are a video measurement agent. From the egocentric video, estimate the approximate distance traveled in meters by the dominant actor or camera. Use global motion and scene-scale cues, and ignore small jitter or in-place head movements. If there is essentially no movement, output \textbf{0}. Answer with a single real-valued number in meters and nothing else.}
        & \textit{Current video clip} \\
    \end{tabular}
    \vspace{-.3cm}
    \caption{\textbf{Question Templates for tasks in EgoTL-Bench.} We replace the \textcolor{red}{highlighted} part in the question template from scene to scene to construct our benchmark.}
    \vspace{-.5cm}
    \label{tab:tasks}
\end{table*}

\subsection{Prompt Templates for the Six Tasks}
\label{sec:supp-prompts}
For completeness, we list the generic prompt schemas used in all experiments.

\subsubsection{Memory-Grounded Task Planning}
For the memory-bank evaluation, we prompt the VLMs with the abstract task description, the corresponding memory-bank video, and four candidate chains-of-thought (options A–D). The model is instructed that the options are shuffled and that it must read and compare all four options. We then ask the model to assign each option a real-valued score between 0 and 1, indicating how well that option matches the given task and memory-bank video, and to output only a JSON object of the form
{ ``scores'': { ``A '': 0.0,  ``B '': 0.0,  ``C'': 0.0, ``D'': 0.0}, ``best'': ``A''}.
Here, "scores" is used to encourage the model to perform fine-grained, relative comparison across all options, while "best" denotes the single selected option (A/B/C/D). During evaluation, we parse the "best" field from the JSON output and compare the corresponding option against the ground-truth CoT label to compute accuracy; the per-option scores are treated as auxiliary signals and are not used directly in the metric.
\subsubsection{Action Reasoning under Complex Environment}

For the action reasoning task, we probe whether VLMs can choose scene-aware actions in cluttered, physically constrained environments.
Each instance consists of an abstract task description, a short egocentric video clip showing the current state of the environment (including obstacles and free space), and four candidate reasoning steps (options A-D) that describe what the camera wearer should do next.
The candidates are constructed so that only one option is globally consistent with both the task and the physical layout (e.g., avoiding blocked paths or unreachable objects), while the others either ignore obstacles, violate basic physical constraints, or are task-irrelevant.

\subsubsection{Next-Action Prediction }
Prompt for next-action prediction.
For the next-action evaluation, we condition the VLMs on the current egocentric video clip and four candidate next-action descriptions. 
The model is instructed that it is an egocentric next-action predictor, that it will see the current clip, and that it must choose exactly one next action from the candidates. 
The prompt explicitly asks the model to output only the chosen option text (verbatim), without any additional explanation or formatting. 
During evaluation, we parse the model's response by matching the returned text against the four candidates and treat the matched candidate as the predicted next action. 
Accuracy is computed by comparing this predicted option with the ground-truth next-action label derived from the original (unshuffled) CSV, while any extra text beyond the selected option is ignored.

\subsubsection{Action Recognition }
For the action-recognition benchmark, we evaluate the VLMs using short egocentric video clips paired with four candidate action descriptions. 
Each candidate is labeled with a letter (A, B, C, or D), and the model is prompted that it is an egocentric video action classifier. 
The prompt presents the current video clip together with the four labeled options and instructs the model to select exactly one option that best matches the action shown in the video and to answer with only a single capital letter (A, B, C, or D), without any additional text.

Given the model's text output, we parse the first valid capital letter in \{A, B, C, D\} and treat it as the predicted label. 
The ground-truth action for each clip is obtained from a separate CSV file and mapped to one of the four options by string normalization and exact matching; this defines the gold letter (A-D). 
We then compare the model's predicted letter with the gold letter to compute accuracy. 
If the model output does not contain any valid letter, the sample is recorded but excluded from the scored set.

\subsubsection{Direction Recognition }

For direction recognition, we evaluate the VLMs on short egocentric video clips labeled with one of five motion directions: \emph{Turn left}, \emph{Turn right}, \emph{Move forward}, \emph{Going up}, or \emph{Going down}. 
The model is prompted as a video grounding agent and instructed to pick the dominant motion direction from this set: it must rely on the global camera/actor motion and explicitly ignore small jitter or head movements. 
The prompt lists the five options in natural language and asks the model to output only the \emph{text} of the chosen option (e.g., ``Turn left'').

At evaluation time, we normalize the model's free-form text response by lowercasing, stripping punctuation, and mapping common paraphrases (e.g., ``go forward'', ``move forward'') to a standard label using a keyword table. 
We then compare this prediction with the ground-truth direction for each clip to compute overall accuracy as well as per-direction accuracy.

\subsubsection{Distance Estimation}
For distance estimation, we ask the VLMs to infer how far the camera (or dominant actor) has moved in each egocentric clip, measured in meters. 
The model is prompted as a video measurement agent and instructed to estimate the approximate traveled distance using global motion and scene-scale cues while ignoring small jitter or in-place head movements; if the scene is essentially stationary, it should output $0$. 
The prompt explicitly requires the model to respond with a single number in meters and no additional text.

For each video, we obtain a scalar ground-truth distance from pre-computed annotations stored in text files and parse the model's response by extracting the first floating-point number as the predicted distance. 
We then compute the absolute error, the relative error, and a mean relative accuracy (MRA) score \cite{yang2024thinking}: for a set of thresholds $\{\theta\}$ in $[0.5, 0.95]$, we check whether the relative error is below $1-\theta$ and average the resulting binary indicators across thresholds. 
This MRA metric rewards predictions that stay consistently close to the ground-truth distance under progressively stricter tolerance levels.

\section{Closed-Source Benchmark Setup}
\label{sec:supp-closed-source}
We evaluate several closed-source VLMs, including GPT-5 \cite{achiam2023gpt}, GPT-4o \cite{hurst2024gpt}, Gemini 2.0 Flash \cite{team2023gemini}, and Gemini 2.5 Flash \cite{comanici2025gemini}, on our EgoTL-Bench and compare them with open-source baselines. On high-level tasks such as memory-conditioned planning and scene-aware action reasoning, all models remain far below human performance, indicating that long-horizon planning is still challenging. Within this low absolute regime, closed-source systems consistently achieve higher scores than open-source models, suggesting a relative advantage in high-level reasoning. At the perceptual layer, however, this gap narrows or even reverses: on next-action prediction and action recognition, strong open-source models are often comparable to, or slightly stronger than, closed-source ones. For direction recognition, both open-source and closed-source models exhibit a strong bias toward predicting ``Move forward'', which makes it difficult for them to reliably distinguish turning motions. Moreover, almost all models perform poorly on distance estimation, indicating that current VLMs still lack robust egocentric distance understanding. Overall, these results show that current VLMs remain far from human-level egocentric spatial understanding and long-horizon reasoning, and that substantial progress is still required.

\section{VLM Fine-Tuning Specification}
\label{sec:supp-vlm-finetune}

\subsection{Model Architectures and Checkpoints}
\label{sec:supp-arch}
We fine-tune Qwen2.5-VL-7B-Instruct \cite{qwen2.5}, a 7B-parameter multimodal vision--language transformer with a frozen vision encoder and a language backbone augmented with cross-modal adapters. To adapt the model to EgoTL without overfitting or incurring the full cost of dense fine-tuning, we adopt low-rank adaptation (LoRA) \cite{hu2022lora} on top of the language backbone. Specifically, we insert rank-16 LoRA adapters into all transformer blocks while keeping both the vision tower and the multimodal projector frozen. This design allows the model to specialize to EgoTL's spatial reasoning distribution while preserving the strong general-purpose capabilities of the base checkpoint.

\subsection{Training Data Configuration}
\label{sec:supp-train-data}
We use a disjoint subset of EgoTL to fine-tune our VLM and then use the test set to evaluate the model. The test set contains 100 task videos spanning 15 scenes. After applying the same curation pipeline as in the main benchmark, we obtain 1.2k Q\&A pairs for training.
\subsection{Evaluation}
\label{sec:eval-lora}
We evaluate the fine-tuned model on the same test set described above. Our model surpasses the strongest baseline across all layers and metrics, with particularly notable gains in high-level planning. At the low-level perceptual layer, it also consistently achieves better performance. In particular, for distance estimation, where all current VLMs struggle, our fine-tuned model attains substantially higher mean relative accuracy, nearly doubling the MRA of the best pre-fine-tuning configuration. These improvements suggest that our human-annotated dataset not only provides VLMs with implicit scale calibration, but also offers reliable supervision, demonstrating that carefully collected human data can substantially improve VLM spatial reasoning.
